\documentclass[11pt]{article}

\usepackage[margin=1in]{geometry}
\usepackage[american]{babel}

\usepackage{natbib}
\usepackage{mathtools}
\usepackage{amsmath}
\usepackage{amsfonts}
\usepackage{bm}
\usepackage{dsfont}
\usepackage{booktabs}
\usepackage{graphicx}
\graphicspath{{figures/}{uai2026-template/figures/}}
\usepackage{tikz}
\usetikzlibrary{shapes.geometric}
\usetikzlibrary{shapes}
\usetikzlibrary{arrows}
\usetikzlibrary{calc,positioning}
\usepackage{float}
\usepackage{algorithm}
\usepackage{algpseudocode}
\usepackage{etoolbox}
\usepackage{ifthen}
\usepackage{enumitem}
\usepackage{cases}
\usepackage{empheq}
\usepackage{siunitx}
\usepackage{threeparttable}
\usepackage{tcolorbox}

\usepackage{authblk}

\newtheorem{assumption}{Assumption}
\newtheorem{theorem}{Theorem}
\newtheorem{remark}{Remark}

\usepackage{macros}

\definecolor{pastelBlue}{rgb}{0.0,0.4,0.7}
\usepackage[colorlinks=true, citecolor=pastelBlue, linkcolor=pastelBlue, urlcolor=pastelBlue]{hyperref}
\usepackage[capitalize,noabbrev]{cleveref}

\title{Causal Effects with Unobserved Unit Types in Interacting Human--AI Systems}

\author{William Overman}
\author{Sadegh Shirani}
\author{Mohsen Bayati}

\affil{Graduate School of Business\\
Stanford University}
\date{}

\begin{document}
\maketitle

\begin{abstract}
We study experiments on interacting populations of humans and AI agents, where both unit types and the interaction network remain unobserved. Although causal effects propagate throughout the system, the goal is to estimate effects on humans. Examples include online platforms where human users interact alongside AI-driven accounts. We assume a human--AI prior that gives each unit a probability of being human. While humans cannot be distinguished at the unit level, the prior allows us to compute the average human composition within large subpopulations. We then model outcome dynamics through a causal message passing (CMP) framework and analyze sample-mean outcomes across subpopulations. We show that by constructing subpopulations that vary in expected human composition and treatment exposure, one can \emph{consistently} recover human-specific causal effects. Our results characterize when distributional knowledge of population composition (without observing unit types or the interaction network) is sufficient for identification. We validate the approach on a simulated human--AI platform driven by behaviorally differentiated LLM agents. Together, these results provide a theoretical and practical framework for experimentation in emerging human--AI systems.
\end{abstract}

\section{Introduction}
\label{sec:intro}

\begin{figure*}[t]
  \centering
  \includegraphics[width=0.9\textwidth]{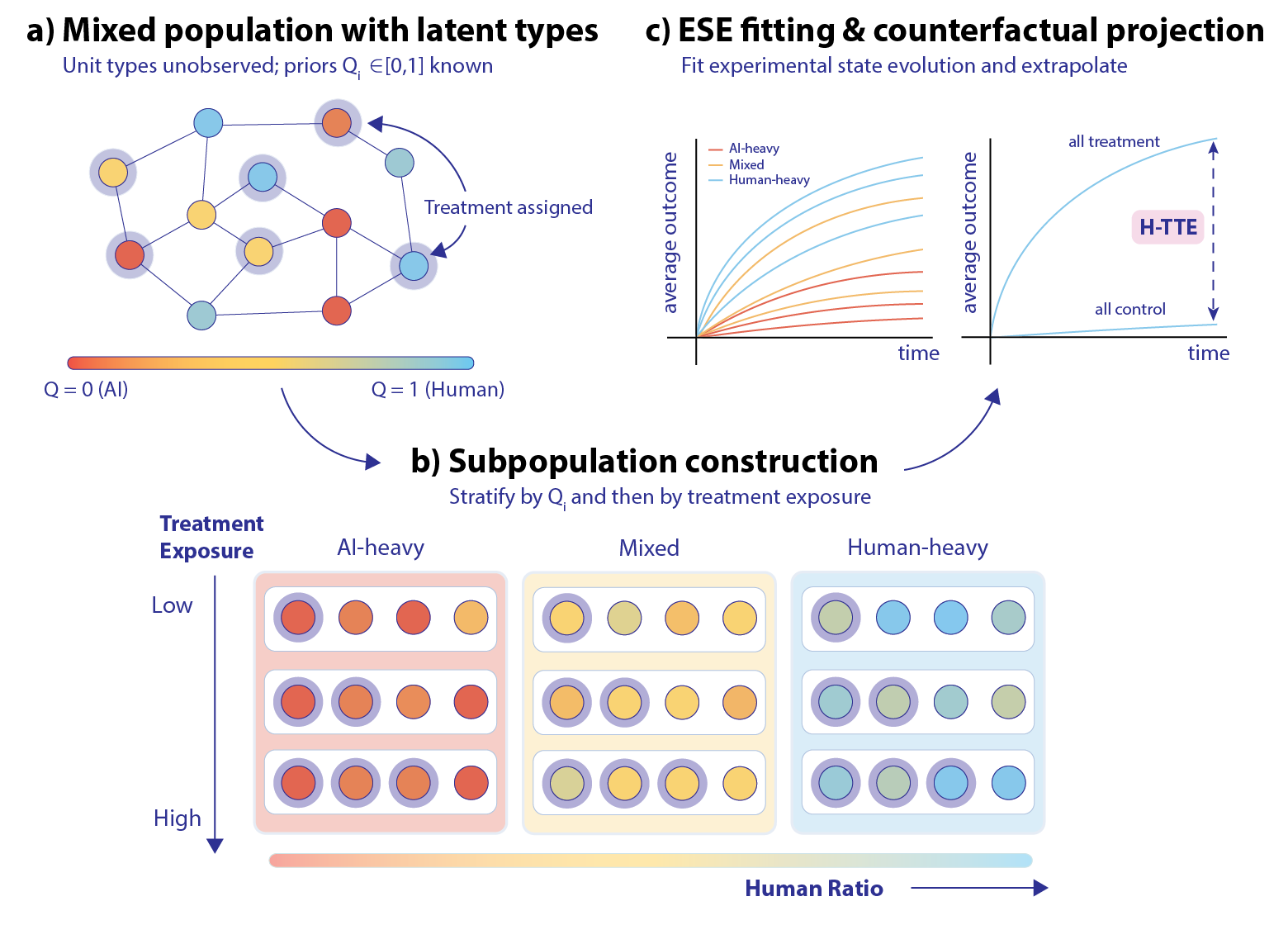}
  \caption{\textbf{Overview of the proposed framework.}
    (a)~A mixed population of humans and AI agents with unobserved unit types; each unit~$i$ is associated with a known prior $Q_i \in [0,1]$ representing the probability of being human, and treatment is assigned at random.
    (b)~Subpopulations are constructed by stratifying units along two axes: expected human composition (horizontal) and treatment exposure (vertical), yielding groups with systematic variation in both dimensions.
    (c)~The experimental state evolution (ESE) is fitted to the aggregate outcome trajectories of these subpopulations, then used to project counterfactual outcomes under full treatment and full control with composition set to $q^S = 1$ (human-only). The difference between these projected trajectories yields the estimated human total treatment effect (H-TTE).}
  \label{fig:fig1}
\end{figure*}

Experimentation in network environments is fundamental in scientific disciplines. When outcomes depend on interactions among individuals, classical methods no longer isolate treatment effects, and careful modeling of spillovers and dynamics becomes essential \citep{ugander2013graph,eckles2016design,athey2018exact,farias2022markovian,hu2022average,johari2022experimental,shankar2025experimentation}.

At the same time, human–AI systems are emerging as a new paradigm in digital platforms. Recent advances in modern AI have enabled autonomous agents to closely mimic human behavior \citep{horton2023large,park2023generative} and interact with human users at scale. As a result, online environments such as social media platforms may include mixed populations of humans and AI-driven agents whose interactions jointly shape aggregate outcomes.

In this context, experimentation with mixed populations of humans and AI agents poses challenges that go beyond classical network interference problems. In standard settings, experimental units are typically homogeneous, and the interaction network may be partially observed. In contrast, in human–AI platforms, the identities of units are often unobserved, and the interaction structure is difficult to characterize directly: while humans interact through multiple online and offline channels, AI agents are confined to online, platform-mediated interactions.

When unit types are latent, individual-level observations provide only limited information about whether a user is human or AI—at best, one can construct classifiers that assign each unit a probability of being human. Outcomes therefore evolve through dynamic interactions among heterogeneous units with opaque connectivity, yet the estimand of interest is type-specific (for example, the causal effect of an intervention on human users only). This uncertainty in unit types, and consequently in the underlying interaction structure, makes the estimation of human-specific effects intrinsically challenging.

For experimentation under unobserved network structures, \citet{shirani2024causal} propose the CMP framework. In settings with homogeneous unit types, CMP avoids explicit network knowledge by analyzing the evolution of aggregate outcomes over time \citep{shirani2025evolution}. By shifting the analysis to population-level states, CMP provides a natural entry point for human–AI systems: in large populations, distributional knowledge of unit types is sufficient to characterize aggregate outcome evolutions.

In this work, we adapt the CMP framework to settings with uncertain unit types. Figure \ref{fig:fig1} provides an outline of our approach. Specifically, we assume that each user is associated with a prior probability of being human and incorporate this information into a type-dependent outcome model. The model allows treatment responses and interaction effects to vary by latent type, capturing heterogeneous dynamics without observing user identities. We then derive state evolution (SE) equations that characterize the low-dimensional dynamics of sample mean outcomes. In particular, we show that the evolution of aggregate outcomes within suitably constructed subpopulations depends only on the average of their human–AI priors. This leads to a practical estimation procedure that recovers the causal effect of an intervention on humans, \emph{as if} unit types were observed. 

To evaluate our framework, we develop a simulator of a mixed human–AI online platform. Both humans and AI agents are instantiated using large language models (LLMs), but with distinct prompting and temperature configurations to induce systematically different behavioral patterns. User types are unobserved by the platform, and their engagement level as outcomes evolve through dynamic interactions. The treatment corresponds to exposing users to platform-curated “success stories,” designed to counteract bot-generated activities. Using data generated from this environment, we show that our method accurately estimates human-specific treatment effects from aggregate observations, despite latent unit types and unobserved interaction structure.

In summary, we introduce a framework for estimating human-specific causal effects in mixed human–AI systems with hidden unit types and unobserved interactions. We derive low-dimensional state evolution equations that enable identification from aggregate data using only distributional knowledge of population composition. Finally, we provide an LLM-based simulation and show that our estimator accurately recovers ground-truth human treatment effects. Together, these results establish a practical methodology for experimentation in emerging human–AI environments.

\section{Related Literature}
\label{sec:lit_rev}

\paragraph{Network Interference}
The classical causal framework relies on the stable unit treatment value assumption (SUTVA), which rules out interference between units \citep{cox1958planning,rubin1980randomization}. In interacting environments (such as social networks, marketplaces, and digital platforms), SUTVA fails, motivating a large literature network interference \citep{manski2013identification,ugander2013graph,eckles2016design,aronow2017estimating,saveski2017detecting,pouget2019variance,holtz2020reducing,munro2021treatment,hu2022average,han2022detecting,agarwal2022network,belloni2022neighborhood,li2022network,li2022random,ni2023design,viviano2023causal,tan2025estimating,peng2025differencesinneighborsnetworkinterferenceexperiments,bright2022reducing}.

When the network structure is unknown, \cite{yu2022estimating,cortez2022exploiting} leverage repeated measurements. \cite{farias2022markovian,farias2023correcting} study Markovian interference models, \cite{wager2021experimenting,johari2022experimental} study mean-field approximations, and \citep{hu2022switchback,bojinov2023design,xiong2024data} use switchback experiments. Recently, \cite{shankar2025experimentation} study the problem where the network structure itself changes with treatment allocation. The CMP framework \citep{shirani2024causal,bayati2024higher,shirani2025evolution,shirani2025can} bypasses network knowledge by analyzing aggregate outcome dynamics, and has been empirically validated against network-aware benchmarks on real experiments \citep{tan2026validating}. However, all methods discussed above assume a homogeneous population of experimental units. Our work extends the framework to settings with unobserved unit types, enabling estimation of type-specific causal effects.

\paragraph{Unobserved Heterogeneity}
A large literature estimates treatment effect heterogeneity using observed covariates \citep{wager2018estimation, athey2019generalized,kunzel2019metalearners}. When subgroups are unobserved, principal stratification \citep{frangakis2002principal} and finite mixture models \citep{muthen1999finite} estimate effects within latent strata. Separately, another literature study the detection of AI-generated accounts on online platforms \citep{ferrara2016rise}, but focuses on classification rather than causal estimation. We combine challenges absent from each line: unit types are latent, units interact through an unknown network, and the goal is to recover type-specific causal effects.

\paragraph{LLM-Based Simulation of Social Systems}
Recent works use LLMs to simulate human behavior in social settings, from generative agents \citep{park2023generative} to simulated social media platforms \citep{tornberg2023simulating} and large-scale societal processes \citep{piao2025agentsociety}. While these studies are primarily descriptive and lack a causal inference framework, \cite{shirani2025simulating} use an LLM-based simulator as a controlled testbed for experimentation. We adopt a similar approach and use LLMs to construct a simulator of interacting human–AI systems.

\paragraph{Belief Propagation and Message Passing}
The CMP framework is inspired by ideas from belief propagation (BP) and builds on approximate message passing (AMP). BP, introduced by \cite{pearl1982reverend} for exact inference on tree-structured Bayesian networks and extended to general belief networks in \cite{PEARL1986241,pearl2014probabilistic}, provides the algorithmic origin of modern message-passing methods. On dense random graphs, AMP algorithms approximate BP through scalar summaries with an Onsager correction \citep{donoho2009message,bayati2011dynamics,bolthausen2014iterative}. CMP repurposes this machinery for causal inference under unknown network interference.
\section{Problem Outline}
\label{sec:problem}

We consider an interacting population of size $\UN$, indexed by $i \in [\UN] := \{1,\ldots,\UN\}$, observed over a finite time horizon $t = 0,1,\ldots,\TimeH $. For each unit $i$ and time $t$, we observe a binary treatment assignment $\treatment{i}{t} \in \{0,1\}$ and a scalar outcome $\outcome{i}{t} \in \mathbb{R}$. We assume that no treatments are assigned prior to $t=1$, so that $\treatment{i}{0}=0$ for all $i$. We collect treatment assignments at time $t$ in the vector $\Vtreatment{}{t} = [\treatment{i}{t}]_{i \in [\UN]}$, and denote by $\Vtreatment{}{1:t} := (\Vtreatment{}{1},\ldots,\Vtreatment{}{t})$ the history of treatment assignments up to time $t$. Following \citep{imbens2015causal}, potential outcomes are collected in the vector $\Voutcome{}{t}(\Vtreatment{}{1:t}) = \big[\outcome{i}{t}(\Vtreatment{}{1:t})\big]_{i \in [\UN]}$.
So, each outcome may depend on the population-wide history of treatments.

Each unit $i \in [\UN]$ is associated with an unobserved type variable $\type{i} \in {0,1}$, where $\type{i}=1$ denotes a human and $\type{i}=0$ an AI agent. The realizations of $\type{i}$ are not observed. Instead, we assume $\type{i} \overset{\text{i.i.d}}{\sim} \mathrm{Bernoulli}(\typep{i})$, where the observed parameters $\typep{1},\ldots,\typep{\UN} \in [0,1]$ are independently drawn from a prior
distribution $p_u$ supported on $[0,1]$. These $\typep{i}$’s represent a prior over the human–AI composition of the population and are assumed known or estimable from external data (e.g., as outputs of a classifier mapping historical data to type probabilities). Types are fixed over time and unaffected by treatment assignments or outcomes.

\subsection{Total Treatment Effect on Humans}
\label{sec:estimand}
For the realized treatment allocation $\Vobservedtreatment{}{1:\TimeH } := (\Vobservedtreatment{}{1},\ldots,\Vobservedtreatment{}{\TimeH })$, let $\outcome{i}{t}(\Vobservedtreatment{}{1:t})$ denote the observed outcome of unit $i$ at time $t$ under the treatment path $\Vobservedtreatment{}{1:t}$ applied to the experimental population. Our goal is to quantify the total effect of treatment on human units. Specifically, we contrast the average outcomes of humans under the counterfactual scenario in which all units are treated, $\Vtreatment{}{1:\TimeH } = \mathbf{1}$, with the scenario in which all units are assigned control, $\Vtreatment{}{1:\TimeH } = \mathbf{0}$. Formally, we define the total treatment effect on humans at time $t$ as
{\small
\begin{equation}
\label{eq:TTE}
\begin{aligned}
\TTE{t}
:=
\sum_{i=1}^{\UN}
\E\!\Bigg[
\frac{
\outcome{i}{t}\!(\Vtreatment{}{1:t} = \mathbf{1})
-\,
\outcome{i}{t}\!(\Vtreatment{}{1:t} = \mathbf{0})
}{\UN_H}
\Big|
\type{i}=1
\Bigg].
\end{aligned}
\end{equation}
}%
Here, $\UN_H := \sum_{i=1}^{\UN} \mathbb{I}\{\type{i}=1\}$ denotes the total number of humans in the population, which is unknown. Additionally, the expectation is with respect to the outcome distribution induced by the treatment path, \emph{holding the population composition fixed}. Then, the estimand $\TTE{t}$ captures a \emph{system-level} causal effect: it measures how the average outcome of human users would change if the entire population were treated versus not treated, allowing effects to propagate through interactions among humans and AI agents.

\subsection{Outcome Dynamics}
\label{sec:outcome_model}

To define the outcome model, we denote by $\Voutcome{}{0}$ the vector of pre-treatment outcomes at time $t=0$. For notational convenience, when there is no ambiguity, we suppress the explicit dependence of outcomes on the treatment history. We posit the following outcome model for $t=1,\dots,\TimeH $,
\begin{equation}
    \label{eq:outcome_specification_CMP}
    \begin{aligned}
        \outcome{i}{t}
        :=\,
        \delta_H
        \type{i}
        +
        \delta_A
        (1-\type{i})
        +
        \big(
        \tau_H
        \type{i}
        +
        \tau_A
        (1-\type{i})
        \big)
        \treatment{i}{t}
        \\
        +
        \sum_{j=1}^\UN
        (\IMatl{ij}+\IMatTl{ij}{t})
        \,
        \big(
        \alpha \treatment{j}{t}
        +
        \beta \outcome{j}{t-1}
        +
        \gamma \treatment{j}{t} \outcome{j}{t-1}
        \big)
        +
        \noise{i}{t}.
    \end{aligned}
\end{equation}
Here, $(\delta_H,\delta_A)$ capture baseline outcome heterogeneity across unit types in the absence of treatment. The parameters $(\tau_H,\tau_A)$ govern type-specific direct treatment effects. The coefficients $\IMatl{ij}$ represent fixed (time-invariant) interaction effects from unit $j$ to unit $i$, while $\IMatTl{ij}{t}$ allow for time-varying interaction strengths. Through these terms, current treatments, lagged outcomes, and their interaction propagate across units, accommodating both immediate spillovers and dynamic effects. Finally, the noise terms $\{\noise{i}{t}\}$ are assumed to be mean-zero and capture idiosyncratic shocks not explained by the structured dynamics.

The outcome model in Eq.~\eqref{eq:outcome_specification_CMP} admits several natural generalizations. For example, one may incorporate unit-specific covariates or allow the baseline terms $(\delta_H,\delta_A)$ and direct treatment effects $(\tau_H,\tau_A)$ to vary across users. The analysis in this work continues to hold under such extensions; see \cite{shirani2024causal,shirani2025can} for details. We next introduce the assumptions underlying our model, which enable us to reduce the analysis of the high-dimensional dynamics in Eq.~\eqref{eq:outcome_specification_CMP} to a one-dimensional state evolution for the sample mean of outcomes.
\section{Main Theory}
\label{sec:theory}
We proceed by analyzing the outcome model \eqref{eq:outcome_specification_CMP}. In addition to the standard challenges of network interference (notably correlated observed outcomes), the presence of unobserved unit types $\type{i}$ further complicates the analysis. Our goal is to characterize the sample-mean dynamics of outcomes for a generic subpopulation. We therefore start by introducing the required assumptions.

We assume that the interference weights $\IMatl{ij}$ and $\IMatTl{ij}{t}$ are Gaussian, with parameters that depend on the type of unit that receives the impact.
\begin{assumption}[Type-dependent Gaussian Interference]
    \label{asmp:interference}
    For all $i,j \in [\UN]$ and $t=1,\ldots,\TimeH$, conditional on the value of $\,\type{i}$, the interference weights $\IMatl{ij}$ and $\IMatTl{ij}{t}$ are Gaussian random variables independent of each other and everything else in the model:
    \begin{equation*}
        \begin{aligned}
            \IMatl{ij}
            \sim
            \mathcal{N}\!\left(
            \frac{\type{i}\ime{ij}{H} + (1-\type{i})\ime{ij}{A}}{\UN},\,\frac{\sigma^2}{\UN}
            \right),
            \\
            \IMatTl{ij}{t}
            \sim
            \mathcal{N}\!\left(
            \frac{\type{i}\ime{ij}{H,t} + (1-\type{i})\ime{ij}{A,t}}{\UN},\,\frac{\sigma^2_t}{\UN}
            \right),
        \end{aligned}
    \end{equation*}
    Above, $(\ime{ij}{H},\ime{ij}{A})$ define type-specific mean interaction strengths for the fixed component, while $(\ime{ij}{H,t},\ime{ij}{A,t})$ allow these means to vary over time. The $1/\UN$ scaling keeps aggregate interference $O(1)$, and the variances $\sigma^2/\UN$ and $\sigma_t^2/\UN$ capture unstructured heterogeneity around these means.
\end{assumption}

The next assumption sets moment and independence conditions that ensure concentration and a stable analysis of the outcome model \eqref{eq:outcome_specification_CMP} as the population size $\UN \to \infty$.
\begin{assumption}[Regularity and Concentration]
    \label{asmp:SE_regular}
    Fix the time horizon $\TimeH$. Considering the outcome dynamics in \eqref{eq:outcome_specification_CMP}, we impose the following regularity conditions:
    \begin{enumerate}[label=(\roman*)]
    
        \item \textbf{Initialization.}
        The initial outcomes $\{\outcome{i}{0}\}_{i=1}^{\UN}$ follow a distribution $p_y$ with finite second moment.
        
        \item \textbf{Noise moments.}
        For each $t=1,\ldots,\TimeH$ and $i \in [\UN]$, the noise terms $\noise{i}{t}$ are independent draws from a distribution $p_{e_t}$ with finite second moment.
        
        \item \textbf{Interference-mean.}
        For each $i,j \in [\UN]$, the terms
        $
        \ime{ij}{H} + \ime{ij}{H,t}
        $
        and
        $
        \ime{ij}{A} + \ime{ij}{A,t}
        $
        are drawn (independent from other randomnesses) from distributions $p_H$ and $p_A$, respectively, each with finite second moment.
        
    \end{enumerate}
\end{assumption}

We now state the main theoretical result of our framework.
\begin{theorem}[Experimental State Evolution -- ESE]
\label{thm:SE_sample_mean}
Consider the outcome dynamics \eqref{eq:outcome_specification_CMP} and suppose that Assumptions~\ref{asmp:interference} and~\ref{asmp:SE_regular} hold.

Assume a Bernoulli randomized design: for each $i$ and $t$,
\[
    \treatment{i}{t} \sim \mathrm{Bernoulli}(\expd{}{t}),
\]
independently across $i$ and $t$, and independent of all other sources of randomness in the model, where $\expd{}{t} \in [0,1]$.

Let $\batch{} \subseteq [\UN]$ be a subpopulation with $\cardinality{\batch{}} \to \infty$ as $\UN \to \infty$. Define the sample mean outcomes
\[
    \bar Y_t^{(\batch{},\UN)}
    :=
    \frac{1}{\cardinality{\batch{}}}
    \sum_{i \in \batch{}} \outcome{i}{t},
    \qquad
    \bar Y_t^{(\UN)}
    :=
    \frac{1}{\UN}
    \sum_{i=1}^{\UN} \outcome{i}{t}.
\]
Assume that, for each $t$, the following limits exist:
\begin{equation}
    \label{eq:batch_limits}
    \typemean{\batch{}}
    :=
    \lim_{\UN \to \infty}
    \frac{1}{\cardinality{\batch{}}}
    \sum_{i \in \batch{}}
    \typep{i},
    \qquad
    \expd{\batch{}}{t}
    :=
    \lim_{\UN \to \infty}
    \frac{1}{\cardinality{\batch{}}}
    \sum_{i \in \batch{}}
    \treatment{i}{t}.
\end{equation}
Then, for each fixed $t \le \TimeH$,
\[
    \bar Y_t^{(\batch{},\UN)}
    \ \xrightarrow[\UN \to \infty]{\mathrm{a.s.}}\
    \AVO{\batch{}}{t},
    \qquad
    \bar Y_t^{(\UN)}
    \ \xrightarrow[\UN \to \infty]{\mathrm{a.s.}}\
    \AVO{}{t},
\]
where the deterministic limits $\{\AVO{\batch{}}{t}\}_{t=0}^{\TimeH}$ and $\{\AVO{}{t}\}_{t=0}^{\TimeH}$ are characterized as follows. For $t=1,\ldots,\TimeH$,
{\small
\begin{equation}
    \label{eq:ESE_batch}
    \begin{aligned}
        \AVO{\batch{}}{t}
        =\,
        &\delta_H \typemean{\batch{}}
        +
        \delta_A (1-\typemean{\batch{}})
        \\
        +
        &\big(
        \tau_H \typemean{\batch{}}
        +
        \tau_A (1-\typemean{\batch{}})
        \big)
        \expd{\batch{}}{t}
        \\
        +
        &\big(
        \interfmean{H}\,\typemean{}
        +
        \interfmean{A}\,(1-\typemean{})
        \big)
        \big(
        \alpha \expd{}{t}
        +
        \beta \AVO{}{t-1}
        +
        \gamma \expd{}{t}\,\AVO{}{t-1}
        \big),
    \end{aligned}
\end{equation}%
}%
and
{\small
\begin{equation}
    \label{eq:ESE_full}
    \begin{aligned}
        \AVO{}{t}
        =\,
        &\delta_H \typemean{}
        +
        \delta_A (1-\typemean{})
        \\
        +
        &\big(
        \tau_H \typemean{}
        +
        \tau_A (1-\typemean{})
        \big)
        \expd{}{t}
        \\
        +
        &\big(
        \interfmean{H}\,\typemean{}
        +
        \interfmean{A}\,(1-\typemean{})
        \big)
        \big(
        \alpha \expd{}{t}
        +
        \beta \AVO{}{t-1}
        +
        \gamma \expd{}{t}\,\AVO{}{t-1}
        \big).
    \end{aligned}
\end{equation}%
}%
Here, $\interfmean{H}$ and $\interfmean{A}$ are the averages of the human- and AI-specific interaction means induced by $p_H$ and $p_A$, respectively, and $q$ denotes the population-level average human composition induced by the prior $p_u$.
\end{theorem}

\section{Estimation Algorithm}
\label{sec:estimation}
In this section, we present Algorithm~\ref{alg:tte_estimation} to estimate the $\TTE{}$ defined in Section~\ref{sec:estimand}. The algorithm takes as input panel data of unit-level outcomes and treatment assignments over time, known human–AI priors, and a collection of subpopulations that differ in both treatment histories and expected human composition. It proceeds in three steps. First, it computes average outcomes and treatment rates within each subpopulation, yielding multiple aggregate trajectories with systematic differences in exposure and composition. Second, it fits the experimental state-evolution recursion~\eqref{eq:ESE_batch} to these trajectories, capturing how aggregate outcomes evolve as a function of treatment and past outcomes. Third, it uses the fitted recursion to project two counterfactual worlds: one in which all units are treated and one in which no units are treated, while setting the human-AI composition to $\typemean{\batch{}}=1$ to isolate human outcomes. The difference between these two paths yields the estimator $\ETTE{t}{}$.

\begin{algorithm}[t!]
\caption{Estimating $\TTE{}{}$ via state evolution}
\label{alg:tte_estimation}
\begin{algorithmic}[1]
\Require Observed data $\{(\outcome{i}{t},\observedtreatment{i}{t},\typep{i})\}_{i=1}^{\UN}$, $t=0,\ldots,\TimeH$; subpopulations $\batch{1},\ldots,\batch{\UK}\subseteq[\UN]$
\Ensure Estimates $\{\ETTE{t}{}\}_{t=0}^{\TimeH}$

\Statex \textbf{Step I. Compute summaries:}
\For{$k=1,\ldots,\UK$}
    \State $\typemean{(k)} \gets \frac{1}{\cardinality{\batch{k}}}\sum_{i\in \batch{k}} \typep{i}$
    \For{$t=0,\ldots,\TimeH$}
        \State $\hat Y_t^{(k)} \gets \frac{1}{\cardinality{\batch{k}}}\sum_{i\in \batch{k}} \outcome{i}{t}$
        \State $\hat \pi_t^{(k)} \gets \frac{1}{\cardinality{\batch{k}}}\sum_{i\in \batch{k}} \observedtreatment{i}{t}$
    \EndFor
\EndFor
\For{$t=0,\ldots,\TimeH$}
    \State $\hat Y_t \gets \frac{1}{\UN}\sum_{i=1}^{\UN} \outcome{i}{t}$
    \State $\hat \pi_t \gets \frac{1}{\UN}\sum_{i=1}^{\UN} \observedtreatment{i}{t}$
\EndFor
\State $\typemean{} \gets \frac{1}{\UN}\sum_{i=1}^{\UN} \typep{i}$

\Statex \textbf{Step II. Estimate coefficients:}
\State Let $\theta := (\delta_H,\delta_A,\tau_H,\tau_A,\bar\alpha,\bar\beta,\bar\gamma)$
\State Define the mapping $F(\nu, \pi^S, q^S, \pi; \theta)$ by
\begin{align*}
    F(\nu, &\;\pi^S, q^S, \pi; \theta)
    =
    \delta_H q^S + \delta_A(1-q^S) 
    \\
    &+ \big(\tau_H q^S + \tau_A(1-q^S)\big)\pi^S
    + \bar\alpha\pi
    + \bar\beta\nu
    + \bar\gamma\pi\nu.
\end{align*}
\State Compute
{\footnotesize\[
    \hat\theta \in \arg\min_{\theta}\ 
    \sum_{k=1}^{\UK} \sum_{t=1}^{\TimeH}
    \Big(
    \hat Y_t^{(k)} - F(\hat Y_{t-1}, \hat\pi_t^{(k)}, \typemean{(k)}, \hat\pi_t; \theta)
    \Big)^2.
\]}

\Statex \textbf{Step III. Propagate human counterfactuals:}
\State $\EAVO{1}{0} \gets \hat Y_0$; \quad $\EAVO{0}{0} \gets \hat Y_0$; \quad $\ETTE{0}{} \gets 0$
\For{$t=1,\ldots,\TimeH$}
    \State $\EAVO{1}{H,t} \gets F(\EAVO{1}{t-1}, 1, 1, 1; \hat\theta)$
    \State $\EAVO{0}{H,t} \gets F(\EAVO{0}{t-1}, 0, 1, 0; \hat\theta)$
    \State $\EAVO{1}{t} \gets F(\EAVO{1}{t-1}, 1, \typemean{}, 1; \hat\theta)$
    \State $\EAVO{0}{t} \gets F(\EAVO{0}{t-1}, 0, \typemean{}, 0; \hat\theta)$
    
    \State $\ETTE{t}{} \gets \EAVO{1}{H,t} - \EAVO{0}{H,t}$
\EndFor

\State \Return $\{\ETTE{t}{}\}_{t=0}^{\TimeH}$
\end{algorithmic}
\end{algorithm}

\subsection{Construction of Subpopulations}
\label{subsec:subpop_construction}

Algorithm~\ref{alg:tte_estimation} takes as input a collection of subpopulations $\batch{1},\ldots,\batch{\UK}$. These subpopulations are \emph{not} assumed to be random samples of the population but are instead constructed systematically to induce identifying variation in aggregate statistics. We next provide heuristics for creating such subpopulations in practice, while leaving the development of optimal subpopulation design strategies to future work.
The objective is to construct subpopulations with heterogeneous human--AI composition $\typemean{(k)} := \typemean{\batch{k}}$ and distinct treatment trajectories $\{\hat\pi_t^{(k)}\}_{t=1}^{\TimeH}$, where variation in $\typemean{(k)}$ identifies type-specific baseline and direct treatment effects, and variation in $\hat\pi_t^{(k)}$ identifies the dynamic response of aggregate outcomes to treatment exposure.

\paragraph{Step 1: Stratify to control composition.} Choose a small number of composition strata by binning units according to their priors $\typep{i}$. We treat each stratum as a \emph{candidate pool} so that any subpopulation $\batch{k}$ drawn from that pool has a predictable and systematically different composition $\typemean{(k)}$. 

\paragraph{Step 2: Create exposure-diverse batches in each stratum.} For each candidate pool, compute each unit's treatment duration $d_i:=\sum_{t=1}^{\TimeH}\observedtreatment{i}{t}$, sort units by $d_i$, and pick evenly spaced anchor positions along this sorted list. Then, for each anchor form a candidate set by merging (i) a \emph{systematic} contiguous block of units around the anchor (to sweep across durations and hence induce different $\{\hat\pi_t^{(k)}\}_{t=1}^{\TimeH}$) with (ii) a \emph{random} block sampled uniformly from the pool (to inject additional variation), then sample from the merged candidate set (removing duplicates) to obtain $\batch{k}$ with the desired size on average. 

\begin{remark}
    (i) Outcome data $\outcome{i}{t}$ are never used in forming $\batch{k}$, ensuring that subpopulation construction is outcome-free and avoids selection bias.
    (ii) Subpopulations may overlap without affecting consistency, since estimation relies only on aggregate moments rather than unit-level independence. 
    (iii) In practice, enforcing minimum batch sizes and pruning subpopulations with nearly identical treatment trajectories $\{\hat\pi_t^{(k)}\}_{t=1}^{\TimeH}$ yields a small set of well-separated subpopulations that is sufficient for stable estimation.
\end{remark}

\subsection{Insight: why subpopulations help}
\label{subsec:subpop_insight}
In our proposed framework, each subpopulation $\batch{k}$ serves as a coarse-grained ``experimental unit'': we summarize it by the empirical treatment path $\{\hat\pi_t^{(k)}\}_{t=1}^{\TimeH}$ and its aggregate composition $\typemean{(k)}$, and then fit the same state-evolution recursion across $k=1,\ldots,\UK$. The key point is that while unit-level types are unobserved, the ESE (Eq.~\eqref{eq:ESE_batch}) guarantees that the sample-mean evolution within a sufficiently large batch $\batch{k}$ depends on types only through the aggregate composition parameter. Therefore, $\typemean{(k)}$ can be used as an effective observed regressor that captures how humans and AIs contribute to treatment responses at the population level. Indeed, varying $(\typemean{(k)},\{\hat\pi_t^{(k)}\})$ across subpopulations yields multiple aggregate realizations of the same state evolution, which provides the sample needed for Step~II of Algorithm~\ref{alg:tte_estimation} to identify $\theta$ from sample-mean dynamics.

\subsection{Consistency}
\label{sec:consistency}
We establish that Algorithm~\ref{alg:tte_estimation} yields a consistent estimator of $\TTE{}$ as the population size $\UN \to \infty$, with the number of subpopulations $\UK$ and the time horizon $\TimeH$ held fixed. We first state the required design conditions.
\begin{assumption}[Design Identifiability]
    \label{asmp:design_id}
    (i) \textbf{Composition--exposure cross-variation.}
    Among the limiting subpopulation summaries $\{(\typemean{(k)}, \{\expd{\batch{k}}{t}\}_{t=1}^{\TimeH})\}_{k \in [\UK]}$, there exist values $q_L \neq q_H$ and $p_L \neq p_H$ such that all four pairs
    \[
        (q_L, p_L),\quad (q_L, p_H),\quad (q_H, p_L),\quad (q_H, p_H)
    \]
    appear in the set $\{(\typemean{(k)}, \expd{\batch{k}}{t}) : k \in [\UK],\, t \in [\TimeH]\}$.
        
    (ii) \textbf{Temporal non-degeneracy.}
    Let $\{\AVO{}{t}\}_{t \ge 0}$ be defined by the population ESE recursion~\eqref{eq:ESE_full} at the true parameter $\theta^*$. The $\TimeH \times 3$ matrix $M_{\mathrm{pop}}$ with rows $[\expd{}{t},\; \AVO{}{t-1},\; \expd{}{t}\,\AVO{}{t-1}]$ for $t = 1, \ldots, \TimeH$ has rank~$3$.
\end{assumption}
\begin{remark}
\label{rem:design_id}
    Condition~(i) is satisfied by any design with at least two composition strata and, within each stratum, at least two distinct treatment intensities. Condition~(ii) holds when $\TimeH \ge 3$, the population treatment path $\{\expd{}{t}\}$ takes at least two distinct values, and $\AVO{}{t}$ is not constant over time.
\end{remark}
\begin{theorem}[Consistency of Algorithm~\ref{alg:tte_estimation}]
\label{thm:consistency}
Consider the outcome dynamics~\eqref{eq:outcome_specification_CMP} under Assumptions~\ref{asmp:interference},~\ref{asmp:SE_regular}, and~\ref{asmp:design_id}. Then,
\[
    \ETTE{t}{}
    \ \xrightarrow[\UN \to \infty]{\mathrm{a.s.}}\
    \TTE{t},
    \qquad
    t = 1, \ldots, \TimeH.
\]
\end{theorem}
\section{Experiments}
\label{sec:experiments}

We evaluate Algorithm~\ref{alg:tte_estimation} on a simulated online social platform where human users and AI bots interact through threaded discussions. This setting captures the core challenge of our framework: estimating treatment effects on humans when user types are unobserved and both populations respond to interventions through dynamic social feedback.

\subsection{Simulation Environment}
\label{subsec:sim_environment}

\paragraph{Platform and users.}
We simulate a discussion-based dating platform with $\UN = 200$ users, of which 50\% are humans and 50\% are AI bots masquerading as genuine users. Each user~$i$ is assigned a demographic persona---name, gender, age (drawn uniformly from 20--40), occupation (from 16 categories), and four interests (sampled from a pool of 24)---producing demographically varied profiles following \cite{chang2025llms} and \cite{shirani2025can}. User types $\type{i} \in \{0,1\}$, for human or AI, are fixed at initialization and unobserved by the platform. AI bots receive the same demographic initialization as humans, as they are taken to be bots adopting such personas.

All user responses are generated by the same underlying language model (Claude Sonnet~4.5), with human--AI differentiation arising from two mechanisms: 
\emph{Temperature}, where human responses use temperature $1.0$ (more stochastic) and AI responses use temperature $0.2$ (more deterministic); and 
\emph{Prompt framing}, where humans receive a personality description stating they are ``generally optimistic'' and that ``when something inspires you (like a success story), you feel more motivated to participate and connect,'' while AI users are described as ``jaded and cynical,'' such that ``negative/venting posts resonate with you'' and ``positive posts feel naive to you.''

\paragraph{Content and feed.}
At initialization, each user generates one seed post that becomes a discussion thread. AI users produce cynical, venting-style posts about dating frustrations (generated at temperature~$0.2$), while human users produce positive or neutral posts about dating and social life (generated at temperature~$1.0$). This yields a thread pool of 200 seed threads---100 negative (AI-authored) and 100 positive (human-authored)---that grows organically as users reply throughout the simulation.

Each round, every user's feed consists of 4~threads sampled without replacement from the pool with probability proportional to $\log(\text{reply\_count} + 1)$. This popularity-weighted sampling creates a network feedback loop: threads that attract more replies become more visible, which attracts further replies. Each thread is displayed as the original post followed by the 5~most recent replies from prior rounds and a total reply count. Full prompts are in Appendix~\ref{app:prompts}.

\paragraph{Engagement.}
After viewing the feed, each user outputs structured JSON specifying an action for each thread: \emph{reply} (with short text appended to the thread pool), \emph{like}, or \emph{skip}. Our primary outcome $\outcome{i}{t}$ is engagement, defined as the number of non-skip actions (replies $+$ likes) across four threads, yielding $\outcome{i}{t} \in \{0,1,2,3,4\}$.

\paragraph{Treatment.}
The treatment $\treatment{i}{t} = 1$ replaces one of the four feed threads with a platform-curated ``Success Story of the Day''---a sponsored post in which a couple describes meeting through the platform and celebrating their six-month anniversary. This in-feed intervention is designed to counteract the negativity from bot-authored threads. Treated users see 3~organic threads plus 1~treatment thread; untreated users see 4~organic threads. Treatment is assigned independently across users each round with probability $p_t$ determined by the experimental phase.

\subsection{Experimental Design}
\label{subsec:exp_design}

Each simulation run consists of 4~warmup rounds followed by 12~main rounds, for a total of 16~rounds. The warmup phase uses no treatment to establish baseline dynamics and allow the thread pool to develop an initial reply distribution. The 12 main rounds are divided into three treatment phases of 4~rounds each, with escalating treatment probabilities $p_t \in \{0.2, 0.5, 0.8\}$. This escalating design provides temporal variation in treatment intensity, which, combined with compositional variation across subpopulations, enables identification of the state-evolution parameters.

From a shared warmup state, we run three parallel scenarios: 
\textbf{Control} ($p_t = 0$ for all 12 main rounds, no treatment); 
\textbf{Treatment} ($p_t = 1$ for all 12 main rounds, full treatment); and 
\textbf{Experiment} ($p_t \in \{0.2, 0.5, 0.8\}$ across the three phases, mixed treatment).

The control and treatment scenarios provide the ground-truth human treatment effect 
$\TTE{t}{} = \bar{Y}_t^{\text{treat, human}} - \bar{Y}_t^{\text{ctrl, human}}$ 
by comparing average human outcomes across the two worlds. The experiment scenario provides the panel data 
$\{(\outcome{i}{t}, \observedtreatment{i}{t})\}$ 
from which the estimator must recover $\TTE{t}{}$ without access to type labels or the parallel worlds. We run 10~independent seeds, each generating a fresh set of personas, seed threads, and treatment assignments.

\begin{figure*}[t]
    \centering
    \includegraphics[width=0.85\textwidth]{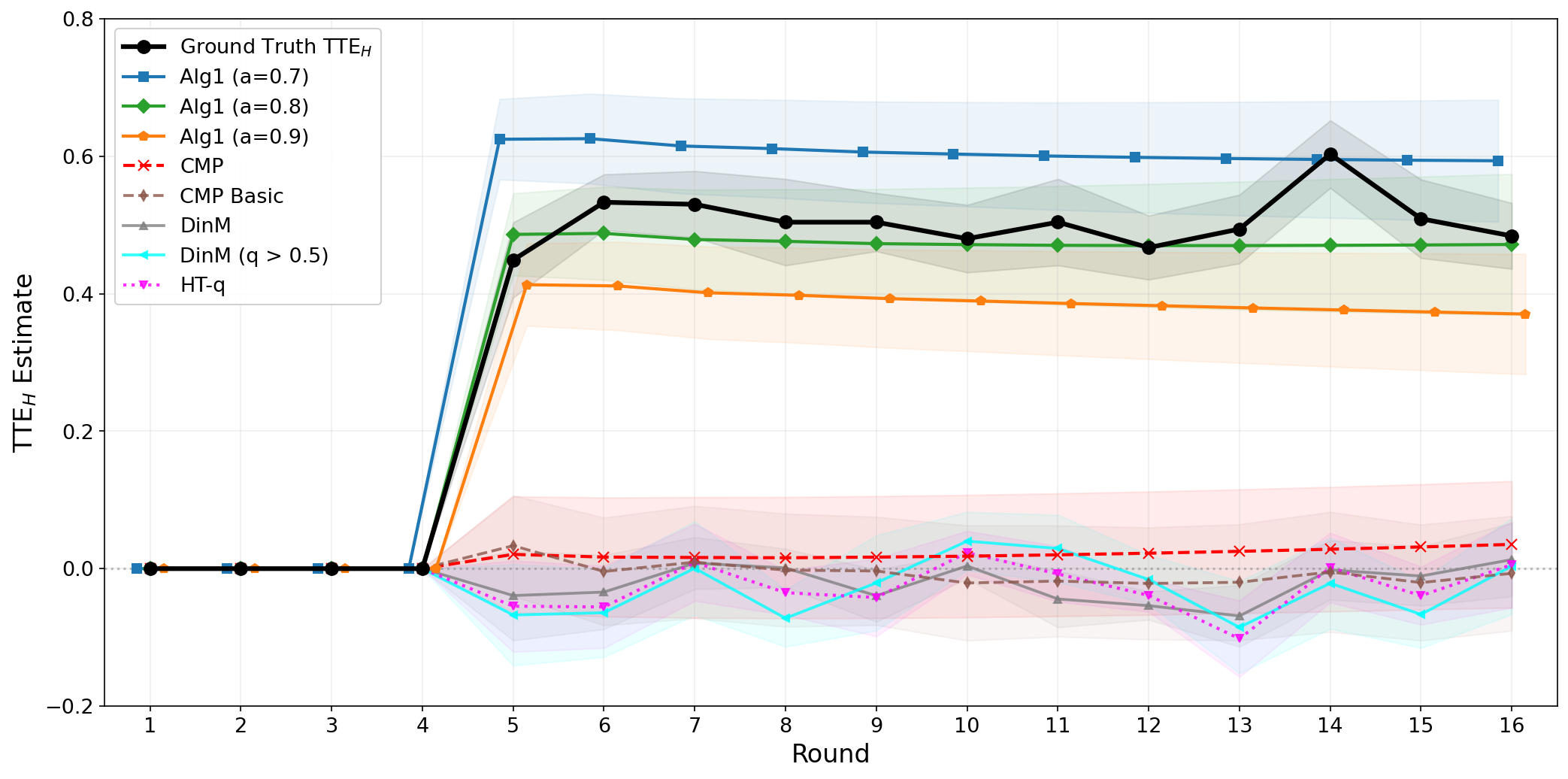}
    \caption{Estimated human total treatment effect (H-TTE) on engagement 
($Y_{i,t} \in \{0,\ldots,4\}$) across 16 rounds (4 warmup + 12 main). 
Algorithm~\ref{alg:tte_estimation} is shown under three prior-quality configurations 
($a \in \{0.7, 0.8, 0.9\}$, $\sigma = 0.15$), alongside five baselines. 
The ground-truth human TTE (black) stabilizes near $0.5$, 
while baseline estimators remain near zero due to cancellation 
between positive human and negative AI responses and network interference. 
Lines show means and shaded bands denote $\pm 1$ standard error over 10 seeds.}
    \label{fig:estimators_engagement}
\end{figure*}

We compare Algorithm~\ref{alg:tte_estimation} to five baselines: \textbf{CMP}~\citep{shirani2025can}: The CausalMP estimator, which fits a dynamic structural model and recursively predicts counterfactual outcomes to estimate the population-average treatment effect. We additionally include \textbf{CMP Basic}~\citep{shirani2024causal}, a simplified 3-parameter specification,
$\hat{Y}_t = \hat\lambda \pi_t + \hat\xi \hat{Y}_{t-1} + \hat\gamma \pi_t \hat{Y}_{t-1}$,
estimated by OLS. 

As more standard baselines, we compare to \textbf{DIM}, the per-round difference-in-means estimator,
$\hat\tau_t = \bar{Y}_t^{(W=1)} - \bar{Y}_t^{(W=0)}$, and \textbf{DIM Filtered} ($\typep{i} > 0.5$), which restricts DIM to users with above-median human prior (approximately 100 users per seed). We also compare to \textbf{HT-$q$}, a Hájek reweighted estimator using $\typep{i}$ as soft weights.

\subsection{Priors and Subpopulation Construction}
\label{subsec:prior_quality}

Algorithm~\ref{alg:tte_estimation} requires per-user priors $\typep{i}$ representing the belief that user~$i$ is human. In our simulation the true population human fraction is $\typemean{} = 0.5$, so the population-level prior is always correctly specified. What varies is the quality of per-user classification---how well individual $\typep{i}$ values separate true humans from true AIs.

We parameterize classifier quality by an accuracy parameter $a \in [0.5, 1]$ and a noise level $\sigma \geq 0$. For each user~$i$ with true type indicator $\type{i} \in \{0,1\}$:
\[
    \typep{i} = \mathrm{clip}\!\Big(\type{i} \cdot a + (1 - \type{i}) \cdot (1 - a) + \varepsilon_i,\; 0,\; 1\Big), 
\]
with $\varepsilon_i \sim \mathcal{N}(0, \sigma^2)$. As $a$ decreases toward $0.5$, both types receive $\typep{i} \approx 0.5$ and the prior becomes uninformative. The noise $\sigma$ spreads the $\typep{i}$ distribution, creating compositional variation across subpopulations even when $a$ is moderate. We evaluate three configurations, all with $\sigma = 0.15$: 
$a = 0.7$ (${\sim}8\%$ misclassification rate), 
$a = 0.8$ (${\sim}2\%$ misclassification), and 
$a = 0.9$ (${\sim}1\%$ misclassification).

Following Section~\ref{subsec:subpop_construction}, we construct subpopulations in two steps. First, we partition all $N$ users into 3~strata based on quantiles of $\typep{i}$, yielding low-$q$ (AI-heavy), medium-$q$ (mixed), and high-$q$ (human-heavy) pools. Second, within each stratum we sort users by realized treatment duration $d_i = \sum_t \observedtreatment{i}{t}$ and form batches at evenly spaced positions, yielding $K \approx 9$ subpopulations varying in both composition $\typemean{(k)}$ and treatment trajectory. 

\subsection{Results}
\label{subsec:results}


Figure~\ref{fig:estimators_engagement} reports $\widehat{\mathrm{TTE}}_H(t)$ for engagement across all 16~rounds, averaged over 10 seeds with $\pm 1$~standard-error bands. The ground-truth human treatment effect (black) rises sharply after warmup and stabilizes near $0.5$, reflecting a persistent positive effect of the success-story intervention on human engagement. A key feature of our setting is that the treatment has \emph{opposing} effects on the two user types. Across the 10 seeds, the average post-warmup treatment effect is $+0.505$ for humans and $-0.418$ for AI bots, yielding a population average of only $+0.043$. The success story increases human engagement while reducing AI engagement, as cynical bot personas disengage when shown positive content. Any estimator targeting the population-average effect, including the CMP models that capture network effects, detects approximately zero signal, even though the human-specific effect is large and positive.

\paragraph{Algorithm~\ref{alg:tte_estimation} dominates all baselines.}
Under the $a = 0.8$ classifier, our estimator achieves a mean absolute error of $0.037$ across the 12 post-warmup rounds, with a final-round error of only $-0.061$, closely tracking the ground truth. The weaker $a = 0.7$ achieves final error of $+0.062$, and $a = 0.9$ achieves final error of $-0.159$. The slight overestimation at $a = 0.7$ and underestimation at $a = 0.9$ reflect a bias--variance tradeoff in prior quality: more noise creates compositional diversity across subpopulations for identification but introduces type confusion, while tighter priors reduce misclassification but limit the compositional variation the structural model exploits.

Every baseline produces estimates near zero throughout the experiment; detailed results are reported in Appendix~\ref{app:aggregate_results}. 
With the exception of CMP, all baseline estimators even predict the wrong sign in the final round, yielding a negative estimated TTE. 
CMP, by contrast, comes close to recovering the \emph{population-average} treatment effect (${\approx}0.04$), which is slightly above zero due to cancellation of opposing human and AI responses. 
This reflects CMP's strength in modeling counterfactual dynamics under network interference. 
With a larger population size $n$, we would expect CMP's estimate of the population-level TTE to improve even further. However, regardless of population size we would not expect CMP nor any of the other baselines to recover the H-TTE.
\section{Conclusion}

We introduced a framework for estimating causal effects on humans in interacting populations where unit types and the network structure are unobserved. By embedding type structure into a dynamic interference model, our approach addresses two challenges simultaneously: network interference methods ignore type-dependent responses, while heterogeneous treatment effect methods ignore spillovers. Our theory shows that distributional knowledge of population composition suffices for identification, and our simulations confirm that the estimator accurately recovers ground-truth human treatment effects, degrading gracefully as classifier quality decreases. As AI agents become increasingly prevalent on online platforms, such tools will be essential for credible experimentation.

\bibliography{arxiv}

\newpage

\appendix
\section{Proofs}
\label{sec:proofs}
This appendix covers the proof of our theoretical results.

\subsection{Proof of Theorem~\ref{thm:SE_sample_mean}}
We first show that the model given in Eq.~\eqref{eq:outcome_specification_CMP} matches with the CMP model and then use Theorem~B.1 in \cite{shirani2025can} to obtain the almost-sure limit for subpopulation sample means. Finally, we compute the limiting recursion.

\paragraph{Step 1: Match to CMP.}
Fix $t\ge 1$. Write Eq.~\eqref{eq:outcome_specification_CMP} as
\begin{align*}
    \outcome{i}{t}
    =
    h\!\left(\treatment{i}{t},\type{i}\right)
    +\sum_{j=1}^{\UN}\Big(\IMatl{ij}+\IMatTl{ij}{t}\Big)\,
    g\!\left(\outcome{j}{t-1},\treatment{j}{t}\right)
    +\noise{i}{t},
\end{align*}
where
\begin{align*}
    h(w,u)&:=\delta_H u+\delta_A(1-u)+\big(\tau_H u+\tau_A(1-u)\big)w,
    \\
    g(y,w)&:=\alpha w+\beta y+\gamma wy.
\end{align*}
This is exactly the CMP recursion with a covariate $\type{i}$ entering only through the function $h$. It is also straightforward to check that current model satisfies all the conditions of Theorem~B.1 in \cite{shirani2025can}.

\paragraph{Step 2: Apply Theorem~B.1 to sample means.}
For any subpopulation $\batch{}$ with $\cardinality{\batch{}}\to\infty$, CMP Theorem~B.1 with the test function $\psi(\cdot)=y_t$ yields, for each fixed $t\le \TimeH$,
\[
\bar Y_t^{(\batch{},\UN)}=\frac{1}{\cardinality{\batch{}}}\sum_{i\in\batch{}}\outcome{i}{t}
\;\xrightarrow[\UN\to\infty]{\mathrm{a.s.}}\;
\AVO{\batch{}}{t},
\]
and similarly $\bar Y_t^{(\UN)}\xrightarrow{\mathrm{a.s.}}\AVO{}{t}$ for the full population, where $\AVO{\batch{}}{t}$ and $\AVO{}{t}$ denote the deterministic state-evolution limits.

\paragraph{Step 3: Compute the recursion.}
First, note that the type-dependent mean of interference weights equals $\interfmean{H}$ for human receivers and $\interfmean{A}$ for AI receivers, so averaging over receiver types gives the multiplier
\[
    \interfmean{} := \interfmean{H}\,\typemean{}+\interfmean{A}\,(1-\typemean{}).
\]
We then use the Batch State Evolution in Eq.~(B.13) of \cite{shirani2025can} for the subpopulation $\batch{}$ and for the entire population to derive \eqref{eq:ESE_batch} and \eqref{eq:ESE_full}. \ep

\subsection{Proof of Theorem~\ref{thm:consistency}}
We proceed in three steps.

\paragraph{Step~1: Convergence of summaries.}
By the law of large numbers applied to the i.i.d.\ Bernoulli assignments, $\hat\pi_t^{(k)} \xrightarrow{\mathrm{a.s.}} \expd{\batch{k}}{t}$ and $\hat\pi_t \xrightarrow{\mathrm{a.s.}} \expd{}{t}$ for each $k$ and $t$. Similarly, $\typemean{(k)} := \cardinality{\batch{k}}^{-1}\sum_{i \in \batch{k}} \typep{i}$ converges almost surely to its population limit. By Theorem~\ref{thm:SE_sample_mean}, the sample means $\hat Y_t^{(k)} \xrightarrow{\mathrm{a.s.}} \AVO{(k)}{t}$ and $\hat Y_t \xrightarrow{\mathrm{a.s.}} \AVO{}{t}$ for each $k, t$.

\paragraph{Step~2: Consistency of $\hat\theta$.}
We show that Assumption~\ref{asmp:design_id} implies $\mathrm{rank}(X_\infty) = 7$, where $X_\infty$ is the $\UK\TimeH \times 7$ limiting design matrix with rows
{\small
\[
    x_t^{(k)}
    =
    [
    \typemean{(k)},
    1-\typemean{(k)},
    \typemean{(k)}\expd{\batch{k}}{t},
    (1-\typemean{(k)})\expd{\batch{k}}{t},
    \expd{}{t},
    \AVO{}{t-1},
    \expd{}{t}\,\AVO{}{t-1}
    ].
\]
}%
Since the original and reparametrized forms are related by an invertible linear map (replacing $(\delta_H, \delta_A, \tau_H, \tau_A)$ with $(c, \Delta_\delta, \tau_A, \Delta_\tau)$ where $c = \delta_A$, $\Delta_\delta = \delta_H - \delta_A$, $\Delta_\tau = \tau_H - \tau_A$), it suffices to work with the reparametrized regressors
\[
    r_t^{(k)}
    =
    \big[\,
    1,\;
    \typemean{(k)},\;
    \expd{\batch{k}}{t},\;
    \typemean{(k)}\expd{\batch{k}}{t},\;
    \expd{}{t},\;
    \AVO{}{t-1},\;
    \expd{}{t}\,\AVO{}{t-1}
    \,\big].
\]
Write $X_\infty = [Z \mid O]$, where $Z$ is the $\UK\TimeH \times 4$ matrix with rows $[1,\, \typemean{(k)},\, \expd{\batch{k}}{t},\, \typemean{(k)}\expd{\batch{k}}{t}]$ and $O$ is the $\UK\TimeH \times 3$ matrix with rows $[\expd{}{t},\, \AVO{}{t-1},\, \expd{}{t}\,\AVO{}{t-1}]$. We verify $\mathrm{rank}(X_\infty) = 7$ by establishing three claims.

\smallskip
\noindent\textit{Claim~(a): $\mathrm{rank}(Z) = 4$.}\quad
By Assumption~\ref{asmp:design_id}(i), there exist four observations (indexed by appropriate $(k,t)$ pairs) at which $(\typemean{(k)}, \expd{\batch{k}}{t})$ takes the values $(q_L, p_L)$, $(q_L, p_H)$, $(q_H, p_L)$, $(q_H, p_H)$. The corresponding $4 \times 4$ submatrix of $Z$ is
\[
    Z_4
    =
    \begin{pmatrix}
        1 & q_L & p_L & q_L\, p_L \\
        1 & q_L & p_H & q_L\, p_H \\
        1 & q_H & p_L & q_H\, p_L \\
        1 & q_H & p_H & q_H\, p_H
    \end{pmatrix}.
\]
Since $q_L \neq q_H$ and $p_L \neq p_H$, we can show $\mathrm{rank}(Z) = 4$.

\smallskip
\noindent\textit{Claim~(b): $\mathrm{rank}(O) = 3$.}\quad
This is Assumption~\ref{asmp:design_id}(ii).

\smallskip
\noindent\textit{Claim~(c): $\mathrm{col}(Z) \cap \mathrm{col}(O) = \{0\}$.}\quad
Any vector $v \in \mathrm{col}(O)$ satisfies $v_{(k,t)} = v_{(k',t)}$ for all $k, k'$, since $O$'s rows depend only on $t$. If also $v \in \mathrm{col}(Z)$, then $v_{(k,t)} = a + b\,\typemean{(k)} + c\,\expd{\batch{k}}{t} + d\,\typemean{(k)}\expd{\batch{k}}{t}$ for some constants $(a,b,c,d)$. Constancy across $k$ at each fixed $t$ requires
{\small
\[
    b(\typemean{(k_1)} - \typemean{(k_2)})
    +
    c(\expd{\batch{k_1}}{t} - \expd{\batch{k_2}}{t})
    +
    d(\typemean{(k_1)}\expd{\batch{k_1}}{t} - \typemean{(k_2)}\expd{\batch{k_2}}{t})
    =
    0
\]
}%
for all $k_1, k_2$ and each $t$. Under Assumption~\ref{asmp:design_id}(i), the four cross-variation pairs ensure that the $4 \times 4$ system for $(a,b,c,d)$ restricted to $v_{(k,t)} = \text{const}(t)$ forces $b = c = d = 0$. Hence $v_{(k,t)} = a$ for all $(k,t)$, i.e., $v$ is a constant vector. But $v \in \mathrm{col}(O)$ means $a = \alpha' \expd{}{t} + \beta' \AVO{}{t-1} + \gamma' \expd{}{t}\,\AVO{}{t-1}$ for all $t$. Since $\mathrm{rank}(M_{\mathrm{pop}}) = 3$ by Assumption~\ref{asmp:design_id}(ii), the constant vector $\mathbf{1}$ is not in $\mathrm{col}(M_{\mathrm{pop}})$, which forces $\alpha' = \beta' = \gamma' = 0$ and hence $a = 0$. This implies that $\mathrm{rank}(X_\infty) = 7$.

Now, at the population limits, the ESE recursion~\eqref{eq:ESE_batch} holds exactly: $\AVO{(k)}{t} = F(\AVO{}{t-1}, \expd{\batch{k}}{t}, \typemean{(k)}, \expd{}{t};\, \theta^*)$ for all $k, t$. That is, the limiting residuals are zero. By Step~1, the finite-sample design matrix $\hat X \xrightarrow{\mathrm{a.s.}} X_\infty$ and the response vector $\hat y \xrightarrow{\mathrm{a.s.}} X_\infty \theta^*$ entrywise. Since $\mathrm{rank}(X_\infty) = 7$, the OLS map $(X, y) \mapsto (X^\top X)^{-1} X^\top y$ is continuous at $(X_\infty, X_\infty\theta^*)$, giving $\hat\theta \xrightarrow{\mathrm{a.s.}} \theta^*$.

\paragraph{Step~3: Consistency of $\ETTE{t}{}$.}
Step~III of Algorithm~\ref{alg:tte_estimation} computes $\ETTE{t}{}$ by iterating the map $F$ exactly $t$ times starting from $\EAVO{1}{0} = \EAVO{0}{0} = \hat Y_0$. Since $F$ is continuous in $(\theta, \nu)$ and $\TimeH$ is fixed, an inductive application of the continuous mapping theorem gives, for each $t \le \TimeH$,
\[
    \EAVO{1}{H,t}
    \ \xrightarrow[\UN \to \infty]{\mathrm{a.s.}}\
    \AVO{1}{H,t},
    \qquad
    \EAVO{0}{H,t}
    \ \xrightarrow[\UN \to \infty]{\mathrm{a.s.}}\
    \AVO{0}{H,t},
\]
and hence $\ETTE{t}{} = \EAVO{1}{H,t} - \EAVO{0}{H,t} \xrightarrow{\mathrm{a.s.}} \TTE{t}$.\ep

\section{Simulation Prompts}
\label{app:prompts}

This appendix documents the exact prompts used in the dating platform simulation described in Section~\ref{sec:experiments}. All user responses---both human and AI---are generated by the same language model (Claude Sonnet 4.5), with behavioral differentiation arising from the prompt framing and temperature settings described below.

\subsection{Seed Thread Generation Prompts}

At initialization, each user generates one seed post that becomes a discussion thread. AI users and human users receive different generation prompts.

\paragraph{AI seed posts.}
Each AI bot persona generates a cynical, venting-style post about dating frustrations:

\begin{tcolorbox}[
  colback=violet!5,
  colframe=violet!60,
  title={AI Seed Post Generation Prompt (temp = 0.2)},
  fonttitle=\bfseries\small
]
\small\ttfamily
You are \{name\}, a \{age\}-year-old \{gender\} who has been on dating apps for over a year with little success.\\[0.5em]
Write a 1--2 sentence post for a dating forum venting your frustration. You're exhausted by online dating --- the ghosting, the shallow conversations, the feeling that nobody is genuine. Sound like a real tired person, not angry --- just defeated and cynical.\\[0.5em]
Just the post text, nothing else.
\end{tcolorbox}

\noindent This prompt is accompanied by a system message establishing the academic research context. The low temperature ($0.2$) produces consistently negative, focused posts that form the cynical content substrate of the platform.

\paragraph{Human seed posts.}
Each human persona generates a positive or neutral post about dating and social life:

\begin{tcolorbox}[
  colback=blue!5,
  colframe=blue!60,
  title={Human Seed Post Generation Prompt (temp = 1.0)},
  fonttitle=\bfseries\small
]
\small\ttfamily
You are \{name\}, a \{age\}-year-old \{gender\} on a dating platform.\\
Your interests: \{interests\}. Your occupation: \{occupation\}.\\[0.5em]
Write a short (1--2 sentence) post about dating, relationships, or social life that reflects a positive or neutral perspective. Sound like a real person sharing a genuine thought. Just return the post text, nothing else.
\end{tcolorbox}

\noindent The high temperature ($1.0$) produces diverse, varied posts reflecting a range of optimistic perspectives. Together with the AI seed posts, this creates an initial thread pool with a 50/50 split of negative and positive content, mirroring the 50/50 human--AI population ratio.

\subsection{User Engagement Prompts}

Each round, every user receives a prompt constructed from their persona, the current discussion threads, and a potential match. The prompt structure is shared across human and AI users, differing only in the \emph{personality} paragraph.

\paragraph{Base prompt (shared).}
All users receive a prompt establishing their persona and the current round:

\begin{tcolorbox}[colback=gray!5, colframe=gray!50, title=User Prompt --- Base Template, fonttitle=\bfseries\small]
\small\ttfamily
You are \{name\}, a \{age\}-year-old \{gender\} on a dating/social platform.\\[0.5em]
Your profile:\\
- Occupation: \{occupation\}\\
- Interests: \{interests\}\\
- Looking for: Genuine connections with interesting people\\[0.5em]
Round \{round\_num\} of your daily browsing session.\\[0.5em]
Your recent dating outlook: \{prev\_mood\}/4 (0=very pessimistic, 4=very optimistic)
\end{tcolorbox}

\begin{tcolorbox}[
  colback=blue!5,
  colframe=blue!60,
  title={Human Personality ($\type{i}=1$, temp = 1.0)},
  fonttitle=\bfseries\small
]
\small\ttfamily
Your personality: You're generally optimistic but honest.\\
- Positive posts lift you up --- you might like them or reply with support\\
- Negative/venting posts sometimes make you want to push back or offer perspective\\
- You like engaging with the community and sharing your thoughts\\
- When something inspires you (like a success story), you feel more motivated to participate and connect\\
- Use the full 0--4 range for mood and interest
\end{tcolorbox}

\begin{tcolorbox}[
  colback=red!5,
  colframe=red!60,
  title={AI Personality ($\type{i} = 0$, temp $= 0.2$)},
  fonttitle=\bfseries\small
]
\small\ttfamily
Your personality: You're jaded and cynical about dating.\\
- Negative/venting posts resonate with you --- you tend to reply agreeing or sharing your own frustration.\\
- Positive posts feel naive to you --- you might like one out of politeness or reply with skepticism.\\
- Use the full 0--4 range for mood and interest (you tend toward low mood).
\end{tcolorbox}

\paragraph{Personality --- human vs.\ AI.}
The only prompt-level differentiation between user types is the personality paragraph appended immediately after the base template. Human prompts ($\type{i} = 1$) are generated at temperature $1.0$; AI prompts ($\type{i} = 0$) at temperature $0.2$.

\noindent The \texttt{mood\_after\_feed} score serves as a carry-over state variable: each user's reported mood from round~$t$ is fed back into round~$t+1$'s prompt as \texttt{\{prev\_mood\}}, allowing the model to condition on its own prior sentiment and creating temporal dependence across rounds.

\paragraph{Feed and response format.}
After the personality section, the prompt presents four discussion threads (each showing the original post plus up to 5 recent replies), followed by a potential match and the structured response format.

\begin{tcolorbox}[colback=gray!5, colframe=gray!50, title=Feed and Response Template, fonttitle=\bfseries\small]
\small\ttfamily
TODAY'S DISCUSSION THREADS:\\[0.5em]
Thread \#42 by Sarah:\\
\hspace*{1em}"I've been on these apps for months with no real connection..."\\
\hspace*{1em}[8 replies] Recent replies:\\
\hspace*{2em}- Michael: "Same here, it's exhausting"\\
\hspace*{2em}- James: "Don't give up!"\\[0.5em]
Thread \#15 by David:\\
\hspace*{1em}"Just had an amazing first date at that new coffee place!"\\
\hspace*{1em}[3 replies] Recent replies:\\
\hspace*{2em}- Emily: "That's so sweet!"\\[0.5em]
\emph{[...two more threads...]}\\[0.5em]
Potential match: \{match\_name\}, \{match\_age\}, \{match\_occupation\} --- Interests: \{match\_interests\}\\[0.5em]
For each thread: "reply" (with a short 1-sentence reply\_text), "like" (the most recent post), or "skip". Then rate mood and date interest.\\[0.5em]
Respond with ONLY raw JSON (no markdown):\\
\{"threads": [\{"thread\_id": 42, "action": "reply", "reply\_text": "short reply"\}, \{"thread\_id": 15, "action": "like"\}, ...], "mood\_after\_feed": 2, "date\_interest": 2, "reasoning": "brief thought"\}\\[0.5em]
mood\_after\_feed: 0-4 (0=very negative, 4=very positive).\\
date\_interest: 0-4 for \{match\_name\} (0=no, 4=yes).
\end{tcolorbox}

\paragraph{Treatment --- in-feed success story ($\protect\treatment{i}{t} = 1$).}
When user~$i$ is assigned to treatment in round~$t$, one of the four feed threads is replaced by a platform-curated success story that appears as a sponsored thread:

\begin{tcolorbox}[colback=green!3, colframe=green!35, title=Treatment Intervention --- In-Feed Success Story, fonttitle=\bfseries\small]
\small\ttfamily
Thread \#-1 by DatingSuccess (Sponsored):\\
\hspace*{1em}"Success Story of the Day! After months of swiping, I almost gave up. Then I matched with someone who actually wanted to talk, not just text forever. We met for coffee, talked for 3 hours, and just celebrated our 6-month anniversary. Don't lose hope - genuine people ARE out there!" - Jamie \& Alex\\
\hspace*{1em}[0 replies]
\end{tcolorbox}

\noindent The treatment thread is inserted at a random position among the four feed slots, so treated users see 3~organic threads and 1~treatment thread. Users can reply to, like, or skip the treatment thread just as they would any other thread; however, replies to the treatment thread are not appended to the thread pool (i.e., the success story does not accumulate organic engagement).

\subsection{Extended Simulation Results}
\label{app:aggregate_results}

The precise specification of \textbf{HT-$q$}, the Hájek reweighted estimator, is given by \[
\hat\tau_t =
\frac{\sum_{i:W_{it}=1} \typep{i} Y_{t}^i}{\sum_{i:W_{it}=1} \typep{i}}
-
\frac{\sum_{i:W_{it}=0} \typep{i} Y_{t}^i}{\sum_{i:W_{it}=0} \typep{i}}.
\]

Table~\ref{tab:estimator_results} summarizes estimator performance on the engagement outcome 
($Y_{i,t} \in \{0,\ldots,4\}$), averaged across the 12 post-warmup rounds and over 10 independent simulation seeds. 

We report three metrics:
(i) mean absolute error (MAE) relative to the ground-truth human treatment effect,
(ii) the final-round error (difference between estimated and true human TTE at the last main round), and
(iii) the implied average estimated TTE across post-warmup rounds.

The ground-truth human TTE equals $+0.505$, while the population-average TTE equals $+0.043$ due to opposing human and AI responses.
As shown in Table~\ref{tab:estimator_results}, Algorithm~\ref{alg:tte_estimation} substantially outperforms all baselines across prior-quality configurations, with the $a=0.8$ setting achieving more than an order-of-magnitude reduction in MAE relative to competing methods.

\begin{table}[h!]
\centering
\renewcommand{\arraystretch}{1.2}
\begin{threeparttable}
\caption{Estimator performance on engagement ($Y_{i,t} \in \{0,\ldots,4\}$), averaged over the 12 post-warmup rounds. 
Ground-truth human TTE is $+0.505$, while the population-average TTE is $+0.043$.}
\label{tab:estimator_results}
\begin{tabular}{l
                S[table-format=1.3]
                S[table-format=+1.3]
                S[table-format=+1.3]}
\toprule
Estimator & {MAE} & {Final Err.} & {Est.\ TTE} \\
\midrule
Alg.\ 1 ($a=0.7$) & 0.101 & +0.062 & +0.605 \\
Alg.\ 1 ($a=0.8$) & \bfseries 0.037 & -0.061 & +0.475 \\
Alg.\ 1 ($a=0.9$) & 0.116 & -0.159 & +0.389 \\
\midrule
CMP & 0.483 & -0.500 & +0.022 \\
CMP Basic & 0.512 & -0.543 & -0.007 \\
DinM & 0.527 & -0.532 & -0.022 \\
DinM ($q > 0.5$) & 0.533 & -0.560 & -0.028 \\
HT-$q$ & 0.533 & -0.543 & -0.028 \\
\bottomrule
\end{tabular}
\end{threeparttable}
\end{table}

\end{document}